\documentclass{llncs}
\usepackage{makeidx}  
\usepackage{graphicx}
\usepackage[hidelinks]{hyperref}
\usepackage{tabularx}
\usepackage{multirow}
\usepackage{caption}
\usepackage{subcaption}

\usepackage{bm}
\usepackage{amsmath}
\usepackage{amsfonts}
\usepackage{braket}

\newcommand{\argmax}{\operatornamewithlimits{arg\,max}}
\newcommand{\indfun}{\operatorname{I}}

\begin{document}
  
\frontmatter          

\pagestyle{headings}  

\mainmatter

\title{Comparing concepts of quantum and classical neural network models for image classification task}

\titlerunning{Comparing concepts of quantum and ...}  
\author{Rafa\l{} Potempa \orcidID{\href{https://orcid.org/0000-0002-0813-0606}{0000-0002-0813-0606}} \and Sebastian Porebski \orcidID{\href{https://orcid.org/0000-0001-9926-4265}{0000-0001-9926-4265}}}

\authorrunning{R. Potempa \and S. Porebski} 

\institute{Department of Cybernetics, Nanotechnology and Data Processing, Faculty of Automatic Control, Electronics and Computer Science, Silesian University of Technology, Gliwice, Poland\\
\email{rafal.potem@gmail.com, sebastian.porebski@polsl.pl}}

\maketitle

\begin{abstract}
While quantum architectures are still under development, when available, they will only be able to process quantum data when machine learning algorithms can only process numerical data. Therefore, in the issues of classification or regression, it is necessary to simulate and study quantum systems that will transfer the numerical input data to a quantum form and enable quantum computers to use the available methods of machine learning. This material includes the results of experiments on training and performance of a hybrid quantum-classical neural network developed for the problem of classification of handwritten digits from the MNIST data set. The comparative results of two models: classical and quantum neural networks of a similar number of training parameters, indicate that the quantum network, although its simulation is time-consuming, overcomes the classical network (it has better convergence and achieves higher training and testing accuracy).

\end{abstract}

\section{Introduction} The paper shows that it is possible to unlock the potential of computational methods dedicated to the Turing-Church-Deutsch machines \cite{deutsch1985} although quantum computers are still in the initial phase of their development \cite{preskill2018}. It is still unclear which of the physical implementations of quantum bits, i.e., by electrons \cite{fujisawa2006}, \cite{mosakowski2016}, \cite{pla2012}, atomic nuclei \cite{morton2008}, photons \cite{riedl2012}, \cite{zhou2012}, or various physical effects, e.g., \cite{josephson1974}, \cite{lucatto2019}, would be final. 

The paper shares the knowledge gathered in the research of the quantum neural networks, which is the extension of exploratory results published in \cite{potempa2021}.
The work was performed using the TensorFlow Quantum \cite{broughton2020} library and similarly to \cite{potempa2021}, it was inspired by the ideas presented in \cite{farhi2018} and \cite{gupta2001}, providing the extension of quantum neural network for solution of MNIST classification from \cite{broughton2020} to all the ten digits. Besides, by using appropriate representation techniques, it was possible to keep the input image larger, however, a more efficient way of quantum representation of images, like FRQI \cite{le2011} might have been used, but this particular example may require considering the use of a recurrent network or sampling layers to store the outcomes of the consecutive measurements on the interaction of quantum and classical network parts, hence increasing the complexity of the solution.
Nevertheless, the technique of dividing the image into four parts helped to reduce the computational effort and made it possible to outperform the classical neural network in comparative experiments. Thus the work successfully simulates the learning of a quantum neural network and shows that quantum learning has a great potential for the development of machine learning.

\section{Material and Methods}
The study deals with a popular MNIST Handwritten Digits Database \cite{lecun2010}. The original data set contains images of digits from 0 to 9 represented by matrices of integer values from 0 to 255, of size $28 \times 28$, representing the grayscale color values of the image pixels. Total number of images is $N=70\ 000$. Database authors \cite{lecun2010} divided it into training and testing subsets of size $\frac{6}{7}N$ and $\frac{1}{7}N$, respectively, hence the data set is prepared for hold-out verification.

\subsection{Quantum computing}
\begin{figure}[htbp]
  \centering
  \begin{subfigure}[t]{0.45\textwidth}
    \centering
    \includegraphics[width=0.7\columnwidth]{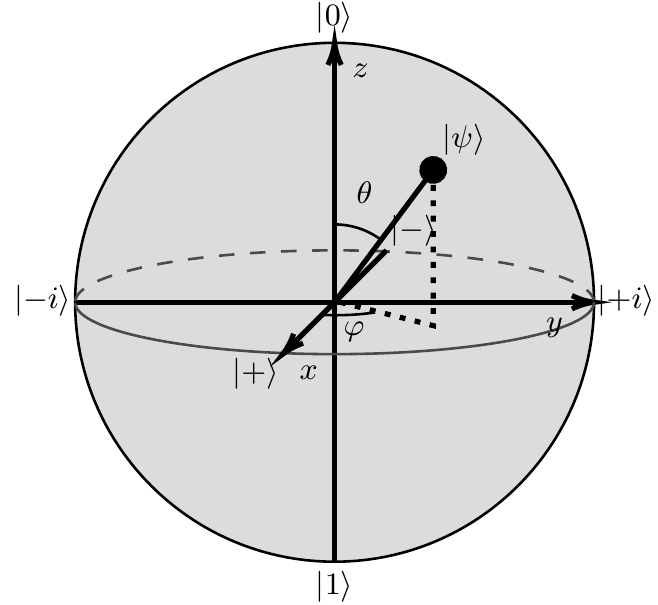}
    \caption{The Bloch sphere}
    \label{fig:bloch_sphere}
  \end{subfigure}
  ~
  \begin{subfigure}[t]{0.45\textwidth}
    \centering
    \includegraphics[width=1\columnwidth]{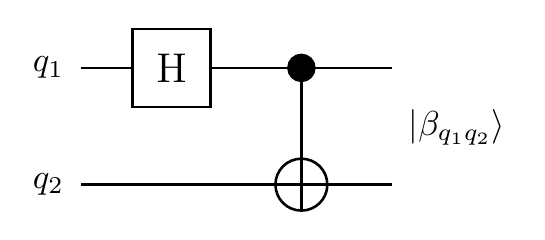}
    \caption{The quantum circuit to introduce a Bell state}
    \label{fig:bell_state_circuit}
  \end{subfigure}
  \caption{Graphical description of quantum data processing basics}
  \label{fig:quantum_processing_basics}
\end{figure}

The quantum bit, or shortly \emph{qubit}, is a conceptual quantum analogy of a classical bit. While a classical bit can be in either of states called $0$ and $1$, a qubit can resemble the quantum analogies of classical states, i.e., $\ket{0}$, $\ket{1}$, or a superposition of those states $\ket{\varphi}$, defined as
\begin{equation}
\label{eq:states}
    \ket{\varphi} = \alpha \ket{0} + \beta \ket{1},
\end{equation}
where $\alpha, \beta \in \mathbb{C}$ and $|\alpha|^2 + |\beta|^2 = 1$ \cite{nielsen2000}. Values $|\alpha|^2$ and $|\beta|^2$ represent the probability of measuring the qubit in state $\ket{\varphi}$ as $\ket{0}$ or $\ket{1}$, respectively, but until the measurement, both states are present and real \cite{nielsen2000}.
Due to values $\alpha$ and $\beta$ being complex numbers, and their combined magnitude being unitary, a qubit can be represented as a vector on a unitary complex sphere, called the Bloch sphere, shown in Fig. \ref{fig:bloch_sphere} and any unitary qubit operation can be treated as a rotation of a unit vector representing state $\ket{\varphi}$, on such sphere \cite{nielsen2000}.

Quantum computing is based on the evolution in time of quantum systems that resemble operations performed on qubits by means of \emph{quantum gates} \cite{deutsch1995}. The gates work not like their classical analogies but rather modify the current position of $\ket{\varphi}$ from (\ref{eq:states}). The result is obtained by initializing all qubits to desired states, application of quantum gates in predefined sequence forming a \emph{quantum circuit} and finally measuring the desired qubits for their internal states. All operations are performed under the control of classical control systems and classical computers, that help to initialize the qubits and preprocess the data into its quantum representation \cite{nielsen2000}.

The most basic quantum gate is the Pauli X gate, i.e., a quantum analogy of NOT gate, that rotates the qubit state by $\pi$ around its $x$ axis and is represented by the Pauli X matrix:
\begin{equation} \label{eq:x-gate}
    \text{X} = \begin{bmatrix}
        0 & 1 \\
        1 & 0
    \end{bmatrix}.
\end{equation}
Another notable example is the Hadamard gate \cite{nielsen2000}, performing a rotation by $\pi$ around $[1, 0, 1]/\sqrt{2}$ axis:
\begin{equation}
    \text{H} = \frac{1}{\sqrt{2}}\begin{bmatrix}
        1 &  1 \\
        1 & -1
    \end{bmatrix},
\end{equation}
which is often used to initialize the qubits into superpositions. The Hadamard gate, combined with a two-qubit controlled-NOT gate (CNOT) \cite{nielsen2000} that flips the target qubit state, only if the control qubit is in $\ket{1}$, as defined in
\begin{equation} \label{eq:cnot}
    \text{CNOT} = \begin{bmatrix} 
        1 & 0 & 0 & 0 \\
        0 & 1 & 0 & 0 \\
        0 & 0 & 0 & 1 \\
        0 & 0 & 1 & 0
    \end{bmatrix},
\end{equation}
can be used to entangle two qubits. To put two qubits $q_1$ and $q_2$ into one of the entangled Bell states $\ket{\beta_{q_1 q_2}}$, $q_1$ is put into a superposition using the Hadamard gate. Next this qubit is used to control the flip of the $q_2$ in the CNOT gate, as shown in Fig. \ref{fig:bell_state_circuit}. This results in the state of $q_2$ being directly dependent on state of $q_1$ and hence entangling it with $q_1$. The determination of a state of one qubit will result in the immediate determination of a state of the other.

The quantum neural network used during the experiments relates to the model proposed in \cite{farhi2018}, combined with classical layers for merge and output. Thus the network forms a hybrid quantum-classical model discussed in \cite{broughton2020}. The presented model is a modification of the quantum neural network proposed in \cite{broughton2020} because it allows classifying all ten MNIST digits, and it is an extension of the author's exploratory experiments reported in \cite{potempa2021}.

\section{Experimental framework}

\subsection{MNIST data set split}
    
In the study, out of the $60\ 000$ images used to tune the network, $1/6$ was dedicated for validation purposes. Let us denote the classification data set as the collection of $N$ pairs of inputs $\boldsymbol{x}_i$ and outputs $y_i$, ($i=1,\,2,\,\dots,\,N$) that is exclusively divided into training ($l$), validation ($v$) and testing ($t$) subsets:
\begin{equation}
    \left\{\boldsymbol{x}^{(l)}_i,y^{(l)}_i\right\}_{i=1}^{N^{(l)}},\quad \left\{\boldsymbol{x}^{(v)}_i,y^{(v)}_i\right\}_{i=1}^{N^{(v)}},\quad \left\{\boldsymbol{x}^{(t)}_i,y^{(t)}_i\right\}_{i=1}^{N^{(t)}}.
\end{equation}
In all experiments $N^{(l)}+N^{(v)}=60\ 000$ and $N^{(t)}=10\ 000$. Let us also denote the general subset
\begin{equation}
\label{eq:data_pair}
  \left\{\boldsymbol{x}^{(*)}_i,y^{(*)}_i\right\}_{i=1}^{N^{(*)}},
\end{equation}
that expresses data pairs without considering their relation to particular subset, but if necessary, general mark ($*$) is replaced with suitable one ($l$, $v$ or $t$).

\subsection{Class label determination}

For a multi-class classification (opposed to the reference approach where only two digits: 3 and 6, are classified \cite{farhi2018}), all ten digits are considered. According to (\ref{eq:data_pair}) each data case $\boldsymbol{x}_i^{(*)}$ is related to its true class label
\begin{equation}
\label{eq:data_label}
  y_i^{(*)} \in \left\{1,\,2,\,\dots,c,\,\dots,C\right\},
\end{equation}
where $C=10$.
Class label (\ref{eq:data_label}) can be also transformed to vector in one-hot encoding form
\begin{equation}
\label{eq:labels_one_hot}
  \boldsymbol{Y}_i^{(*)} = \left[Y_{i,1}^{(*)},\,Y_{i,2}^{(*)},\,\dots,\,Y_{i,c}^{(*)},\,\dots,\,Y_{i,C}^{(*)}\right],
\end{equation}
where $Y_{i,c}^{(*)} = 1$ if $y_i^{(*)} = c$ and $Y_{i,c}^{(*)} = 0$ otherwise, hence only one element of (\ref{eq:labels_one_hot}) is equal to one and the rest is zero.

\subsection{Sub-sampling} \label{scn:sub-sampling}
Each data case in the original data set is a grayscale image of $28\times 28$ pixels. This would require 784 input neurons in the network models. Simulation of a quantum circuit that processes such data dimension is hardly applicable on a classical machine.
To reduce the complexity of simulating the neural network's quantum circuit and to allow operating on an ordinary machine, the images are scaled to the resolution of $8 \times 8$ pixels using the dedicated resize method from \cite{tensorflow2015}.
The resulting images are split into four parts of size $4 \times 4$ pixels, giving the final reduction of memory required to simulate the circuit using a modified version of quantum image representation proposed by \cite{broughton2020} from $2^{784}$ to roughly $4 \cdot 2^{16}$ classical bits, giving the final mapping of
\begin{equation}
\label{eq:data_case_dim}
    B^{28} \times B^{28} \rightarrow [E^4 \times E^4]^4,
\end{equation}
where $B = \{0, 1, ..., 255\}$ and $E = [0, 1]$.

\subsection{Quantum data representation}

\begin{figure}[htbp]
    \centering
    \begin{subfigure}[t]{0.32\textwidth}
        \centering
        \includegraphics{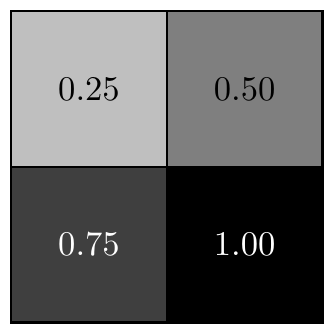}
        \caption{}
    \end{subfigure}
    \hfill
    \begin{subfigure}[t]{0.33\textwidth}
        \centering
        \includegraphics[width=0.65\textwidth]{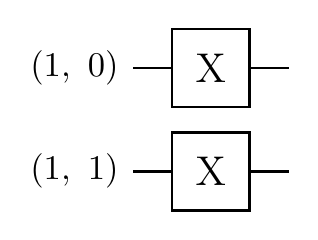}
        \caption{}
    \end{subfigure}
    \hfill
    \begin{subfigure}[t]{0.33\textwidth}
        \centering
        \includegraphics[width=\textwidth]{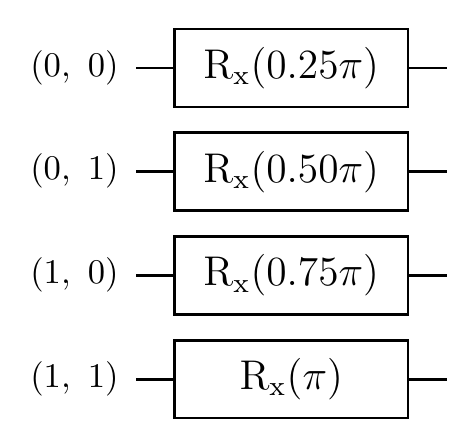}
        \caption{}
        \label{sfig:mnist_encoding_with_Rx}
    \end{subfigure}
    \caption[Quantum encoding of example image]{Quantum encoding of example image (a) with $\text{X}$ gates (b) \cite{broughton2020} and $\text{R}_\text{x}$ gates (c) \cite{potempa2021}}
    \label{fig:quantum_image_encoding}
\end{figure}

The image fragments with the size of $4 \times 4$ are represented using a method similar to \cite{broughton2020}, where the image pixel is correlated directly with a single qubit. The grayscale value determines the rotation of a qubit corresponding to the pixel position. The rotation is performed using the generalization of (\ref{eq:x-gate}), being the $\text{R}_\text{x}$ gate, allowing for rotation around the $x$ axis by arbitrary angle $\theta$, as defined in
\begin{equation} \label{eqn:rotation_Rx_gate}
    \text{R}_\text{x}(\theta) = e^{-i \theta \text{X}/2} = \begin{bmatrix}
          \cos{\frac{\theta}{2}} & -i\sin{\frac{\theta}{2}} \\
        -i\sin{\frac{\theta}{2}} &   \cos{\frac{\theta}{2}}
    \end{bmatrix},
\end{equation}
where $\theta = \pi x$ is the rotation angle obtained by mapping the normalized pixel grayscale value onto ${[0, \pi]}$. Hence the original values between 0 and 255 (from 0.0 to 1.0 after normalization) are represented by $\ket{0}$ and $\ket{1}$ state, respectively. The example is shown in Fig. \ref{sfig:mnist_encoding_with_Rx}.

\subsection{Building a model of a quantum neural network}

\begin{figure}
    \centering
    \includegraphics[width=\textwidth]{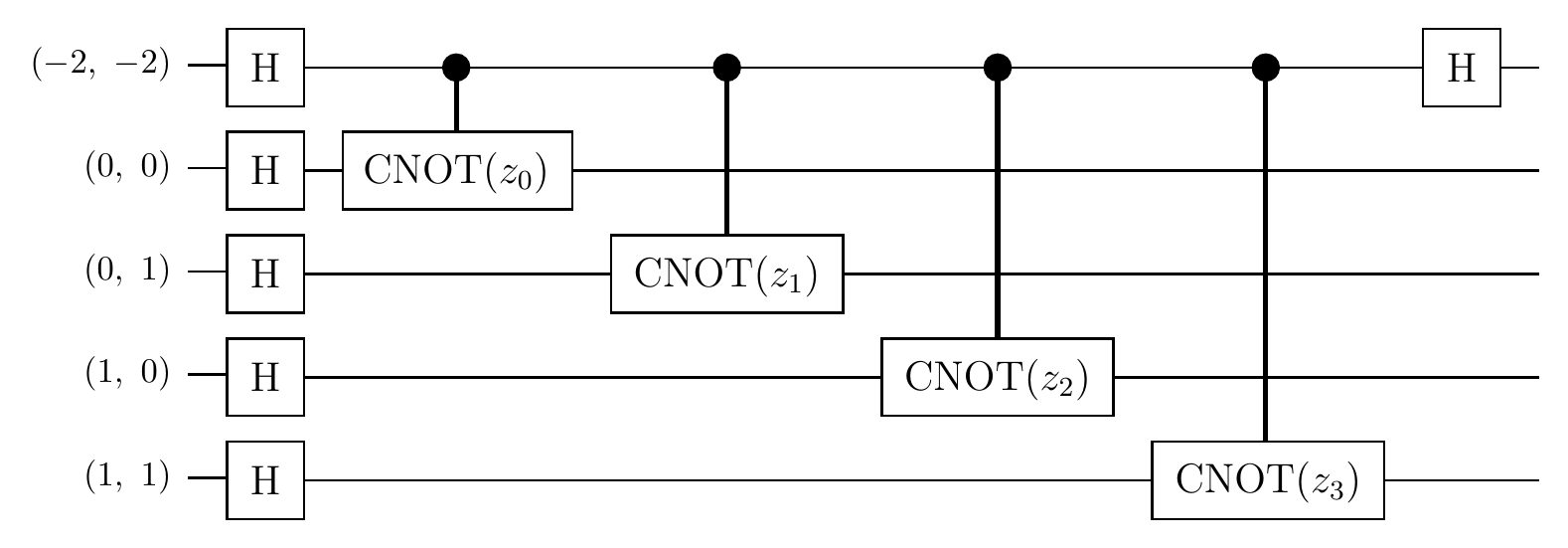}
    \caption{Layout of a simplified version of PQC used for experiments, where $(-2, -2)$ qubit is the readout qubit and the other qubits are used for image pixels representation \cite{potempa2021}.}
    \label{fig:pqc}
\end{figure}

The quantum layer is used as an analogy of classical dense layer and is based on the idea of Parametrized Quantum Circuit (PQC). The layer responsible for insight is a quantum circuit with trainable parameters. The PQC uses parametrized CNOT gates defined as
\begin{equation} \label{eq:cnot_pow_matrix}
    \text{CNOT}(z) = \begin{bmatrix}
        1 & 0 &  0   &  0   \\
        0 & 1 &  0   &  0   \\
        0 & 0 &  gc  & -igs \\
        0 & 0 & -igs &  gc
    \end{bmatrix},
\end{equation}
where $c = \cos(\frac{\pi}{2} z)$, $s = \sin(\frac{\pi}{2} z)$, $g = e^{\frac{i \pi z}{2}}$ \cite{cirq2020}, with $z$ being the layer trainable parameters, i.e., the network learns what should be the scaling of all controlled negation operations. The PQCs are composed of 16 data qubits, that control the single readout qubit. The simplified PQC, that would be used if the images were $2 \times 2$ pixels is shown in Fig. \ref{fig:pqc}. 

\begin{figure}[htbp]
  \centering
  \begin{subfigure}[b]{0.45\textwidth}
    \centering
    \raisebox{0.2\textwidth}{\includegraphics[height=0.6\textwidth]{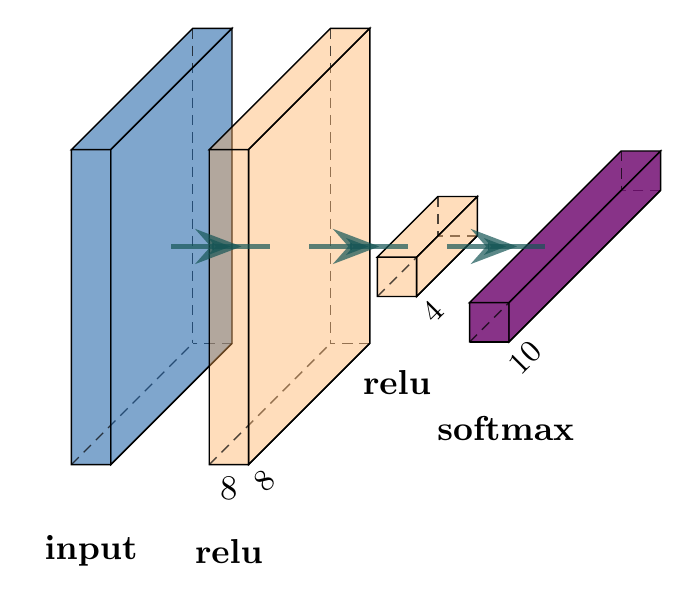}}
    \caption{}
    \label{sfig:model_baseline}
  \end{subfigure}
  \begin{subfigure}[b]{0.45\textwidth}
    \centering
    \includegraphics[height=1\textwidth]{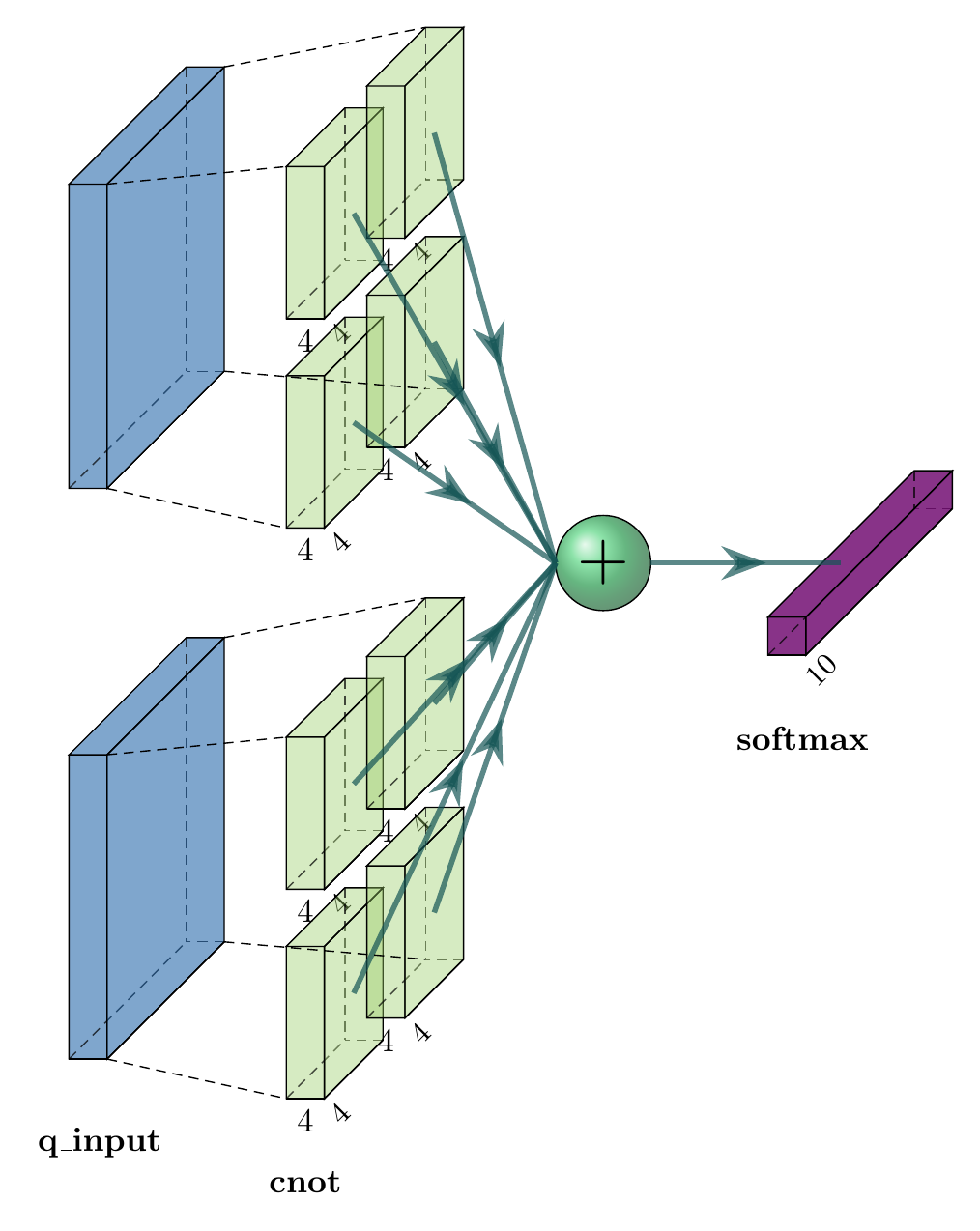}
    \caption{}
    \label{sfig:model_quantum}
  \end{subfigure}
  \caption{Overviews of the classical (a) and quantum neural network (b) model architectures}
  \label{fig:overviews}
\end{figure}

The quantum model is composed of two parallel layers, both constructed of four $4 \times 4$ qubit PQCs. The data is duplicated and fed into the layers in the form of $4 \times 4$ image chunks that form the original, scaled-down image. The output of these two quantum layers is concatenated using the classical merge layer and passed through the softmax layer for the final class membership certainty values prediction, as shown in Fig. \ref{sfig:model_quantum}. The resulting number of trainable parameters is 218.

\subsection{Building a model of classical neural network}

The classical neural network model, compared to the prototype quantum model from the TensorFlow Quantum library \cite{broughton2020}, has a standard structure. Using the TensorFlow library \cite{tensorflow2015}, a network of the form of multi-layer perceptron with one hidden layer was built, ensuring that the number of parameters is similar to the quantum model. Thus, there is an input layer transforming the image to the vector shape, the hidden layer with four neurons, and the output layer with ten neurons. The resulting number of trainable parameters is 310, which is a number higher than that of the parameters of the quantum model, but this is related to the fact that the quantum images required to be split and that PQCs are not interconnected, leading to the reduction in the number of parameters in quantum model. However, the model layouts are the same.

\subsection{Training, validation, and testing}
Both models, quantum and classical neural networks (see Fig. \ref{fig:overviews}), are designed to be $p=64$ input, and $C=10$ outputs classifiers. Hence the quality of the result of training, validation, and testing can be made analogous. A dedicated functions from the TensorFlow libraries \cite{tensorflow2015,broughton2020} are used for training with validation.
Predicted output vector of studied models can be of the following form:
\begin{equation}
\label{eq:pred_one_hot}
  \widehat{\boldsymbol{Y}}_i^{(*)} = \left[\widehat{Y}_{i,1}^{(*)},\,\widehat{Y}_{i,2}^{(*)},\,\dots,\,\widehat{Y}_{i,c}^{(*)},\,\dots,\,\widehat{Y}_{i,C}^{(*)}\right],
\end{equation}
where $\widehat{Y}_{i,c}^{(*)}$ is the value of the $c$th output neuron of the network that indicate membership of the $i$th data case to the $c$th class.
The output result is defined as
\begin{equation}
\label{eq:pred_label}
  \hat{y}_i^{(*)} = \argmax\limits_{1\leq c \leq C}\left[\widehat{\boldsymbol{Y}}_i^{(*)}\right]
\end{equation}
for the purpose of its comparison with (\ref{eq:data_label}).
Implementation of the neural networks required some training and validation presumptions, e.g., about loss calculation and choosing the optimization method, hence the categorical cross-entropy (CCE) was chosen as loss function for the multi-class classifier training and validation. It is calculated for each true class vector ${\boldsymbol{Y}}_i^{(*)}$ (\ref{eq:labels_one_hot}) and predicted output $\widehat{\boldsymbol{Y}}_i^{(*)}$ (\ref{eq:pred_one_hot}), separately, and averaged for the subset:
\begin{equation}
\label{eq:cce}
   \text{CCE}^{(*)} = \frac{-1}{N^{(*)}} \sum\limits_{i=1}^{N^{(*)}}\left[\sum\limits_{c=1}^C Y_{i,c}^{(*)}\cdot \log\widehat{Y}_{i,c}^{(*)}\right].
\end{equation}
The Adam algorithm, which is a stochastic gradient descent method based on adaptive estimation of first- and second-order moments serves for a networks' parameters optimization \cite{tensorflow2015}. Obviously, the network testing involved feeding it with $\boldsymbol{x}^{(t)}_i$ to predict $\hat{y}^{(t)}_i$ and finally, compare these with $y^{(t)}_i$, ($i=1,\,\dots,\,N^{(t)}$). To measure general performance of studied models, accuracy is calculated as
\begin{equation}
  \text{ACC}^{(*)} = \frac{1}{N^{(*)}}\sum\limits_{i=1}^{N^{(*)}}\indfun\left(\hat{y}_i^{(*)}=y_i^{(*)}\right),
  \label{eq:acc}
\end{equation}
where $\indfun(\cdot)$ equals one when the $i$-th data case is correctly classified and zero otherwise. Presented results relate to $\text{ACC}^{(l)}$ and $\text{ACC}^{(v)}$ that are calculated in each epoch (indexed by $k$), hence they are denoted as $\text{ACC}_k^{(l)}$, $\text{ACC}_k^{(v)}$, respectively. The testing subsets does not contain equinumerous classes, hence the second method of evaluation is balanced accuracy 
\begin{equation}
    \text{BA}^{(t)} = \frac{1}{C} \sum\limits_{c=1}^C \left(\frac{\sum\limits_{i=1}^{N^{(t)}}\indfun\left({y}_i^{(t)}=c\wedge\hat{y}_i^{(t)}=c\right)}{\sum\limits_{i=1}^{N^{(t)}}\indfun\left({y}_i^{(t)}=c\right)}\right)
    \label{eq:ba},
\end{equation}
which is the average of $C$ class-precision values.

\subsection{Results}

The simulation of the quantum network is aimed at checking its classification possibilities in comparison with the classical network using a similar architecture of similar size. Naturally, the simulation of a quantum network is highly time-consuming because it requires the numerical equivalent of quantum computations, so the training and validation time cannot be compared. Nevertheless, the change of quantum and classical network performance from epoch to epoch allows their essential comparison. In order to carry out the calculations in a reasonable time, it was chosen to subject both networks to training and validation for ten epochs.

\begin{figure}[p]
    \centering
    \includegraphics[width=0.95\textwidth]{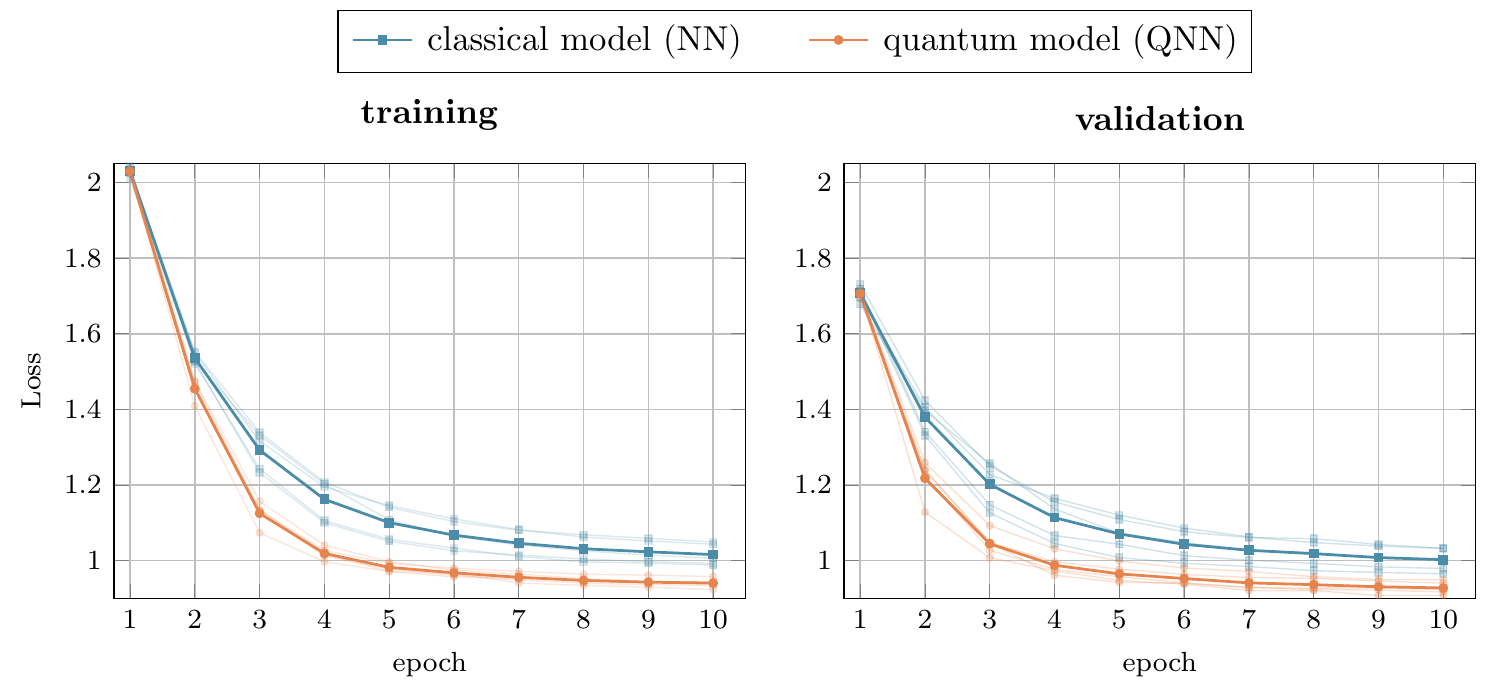}
    \caption{Changes of loss function (\ref{eq:cce}) for classical and quantum models in ten epochs. Transparent lines relates to single experiment and solid lines are mean results.}
    \label{fig:res_loss}
\end{figure}

\begin{figure}[p]
    \centering
    \includegraphics[width=0.95\textwidth]{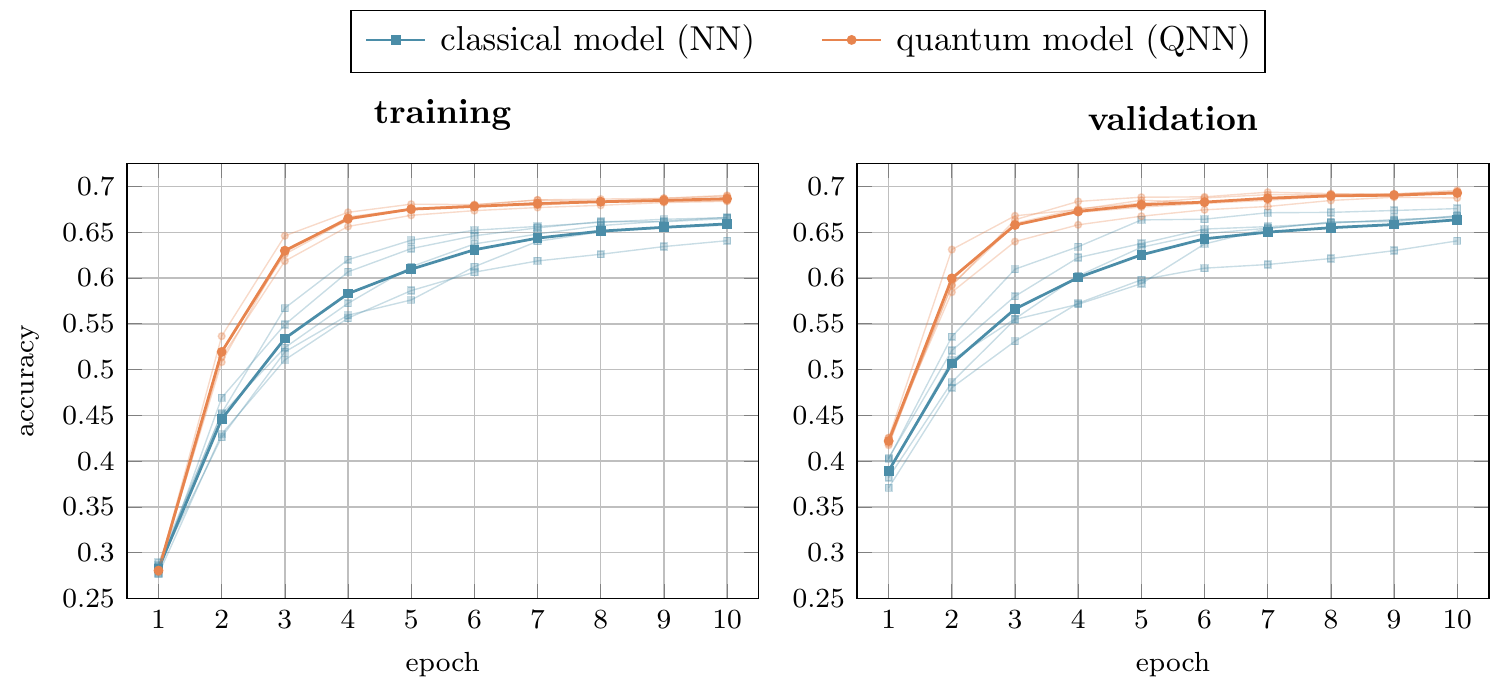}
    \caption{Changes of accuracy (\ref{eq:acc}) for classical and quantum models in ten epochs. Transparent lines relates to single experiment and solid lines are mean results.}
    \label{fig:res_acc}
\end{figure}

The aforementioned training and validation time was enough to check whether the training of the quantum network was successful and to compare its work with the classical network. 
Figures \ref{fig:res_acc} and \ref{fig:res_loss} present how $\text{ACC}_k^{(l)}$, $\text{ACC}_k^{(v)}$ and $\text{CCE}_k^{(l)}$, $\text{CCE}_k^{(v)}$ changed during training and validation of both models. As the result of each of five experiments $\text{ACC}_k^{(t)}$ and $\text{BA}_k^{(t)}$ are calculated.

\begin{table}[p]
    \centering
    \begin{tabularx}{\textwidth}{*8{>{\centering}X}}
    measure & model & $k=1$ & $k=2$ & $k=3$ & $k=4$ & $k=5$ & average \tabularnewline\hline
    \multirow{2}{*}{$\text{ACC}_k^{(t)}$} & classical & 67.71\% & 67.04\% & 67.32\% & 68.43\% & 67.06\% & 67.51\%\tabularnewline
    & quantum & 70.67\% & 70.59\% & 69.72\% & 70.78\% & 70.83\% & 70.52\%\tabularnewline \hline
    \multirow{2}{*}{$\text{BA}_k^{(t)}$} & classical & 67.24\% & 66.56\% & 66.88\% & 67.89\% & 66.52\% & 67.02\%\tabularnewline
    & quantum & 70.15\% & 70.10\% & 69.20\% & 70.32\% & 70.31\% & 70.02\%\tabularnewline\hline
    \end{tabularx}
    \caption{Comparison of $\text{ACC}_k^{(t)}$ (\ref{eq:acc}) and $\text{BA}_k^{(t)}$ (\ref{eq:ba}) of each of five hold-out experiments for classical and quantum model}
    \label{tab:test_results}
\end{table}

According to both Figures (\ref{fig:res_loss} and \ref{fig:res_acc}) one can see that the quantum neural network model (whether considering a change of accuracy and loss for training and validation steps) has a faster convergence, i.e., it achieves a stable result quicker. Greater stability can also be seen by analyzing the results of individual five experiments (transparent lines). Both models tend to minimize the loss function evenly, which cannot be said about the accuracy. For the latter, the classical model has a greater values dispersion for a single epoch, and they do not change as smoothly as for the quantum model. Particularly poor convergence stability of individual experiments can be seen in validation (right diagram in Fig. \ref{fig:res_acc}). The observation is also confirmed in the testing stage of the models. The advantage of the quantum model is proven by the accuracy and balanced accuracy values for each of the five experiments presented in Table \ref{tab:test_results}.

\section{Discussion and Conclusions}
The most important motivation of the research was to propose a reliable quantum model, which would allow achieving results similar to the classical neural network within a reasonable computation time. As it turns out, this can be achieved, and additionally, the quantum model can outperform the classical in two aspects. The first is the convergence of the quantum network learning, which is a much smaller number of iterations before achieving the flattening of the accuracy and loss functions. In this aspect, the classical network seems to be much slower and less stable. The second aspect is that the quantum network, although having a similar architecture, achieved higher training, validation, and testing results for each of the five repeated verification experiments.

Although the advantage of the quantum model over the classical model in this work is still of a simulation nature, it is essential that the proposed algorithm converts numerical data into their quantum representation, i.e., it prepares image data to a form that can be directly sent to and processed by the quantum architecture. Thus, proposing algorithms for processing numbers into quantum representations and trying to solve real classification problems, such as image recognition, is of great importance in the field of quantum machine learning. 

The experimental results provide a reason to develop the topic through further research. Among them lies the simplification of the used quantum image representation method or the representation by FRQI method \cite{le2011}. Another interesting aspect is the serviceability of the quantum models with, e.g., custom layers for automated data preprocessing and the research on the performance of more complex hybrid quantum-classical neural network models, together with the topics of scalability and operation performance of the networks on quantum processing units, that introduce noise related to the physical implementations of qubits and quantum gates.

\end{document}